\documentclass[10pt,twocolumn,letterpaper]{article}

\usepackage{cvpr}              

\usepackage{indentfirst}
\usepackage{stfloats}
\usepackage{multirow}

\definecolor{cvprblue}{rgb}{0.21,0.49,0.74}
\usepackage[pagebackref,breaklinks,colorlinks,allcolors=cvprblue]{hyperref}

\def\paperID{13079} 
\def\confName{CVPR}
\def\confYear{2025}

\title{Tokenize Image Patches: Global Context Fusion for Effective Haze Removal in Large Images}

\author{Jiuchen Chen \qquad \qquad Xinyu Yan \qquad \qquad Qizhi Xu$^{*}$ \qquad \qquad Kaiqi Li\\
Beijing Institute of Technology\\
{\tt\small \{castlechen, yanxinyu, qizhi, kaiqilee\}@bit.edu.cn}
}

\begin{document}
\maketitle

\renewcommand{\thefootnote}{\fnsymbol{footnote}}
 \footnotetext[1]{corresponding author}

 \begin{abstract}

Global contextual information and local detail features are essential for haze removal tasks. Deep learning models perform well on small, low-resolution images, but they encounter difficulties with large, high-resolution ones due to GPU memory limitations. As a compromise, they often resort to image slicing or downsampling. The former diminishes global information, while the latter discards high-frequency details. 
To address these challenges, we propose DehazeXL, a haze removal method that effectively balances global context and local feature extraction, enabling end-to-end modeling of large images on mainstream GPU hardware. Additionally, to evaluate the efficiency of global context utilization in haze removal performance, we design a visual attribution method tailored to the characteristics of haze removal tasks. Finally, recognizing the lack of benchmark datasets for haze removal in large images, we have developed an ultra-high-resolution haze removal dataset (8KDehaze) to support model training and testing. It includes 10000 pairs of clear and hazy remote sensing images, each sized at 8192 $\times$ 8192 pixels. Extensive experiments demonstrate that DehazeXL can infer images up to 10240 $\times$ 10240 pixels with only 21 GB of memory, achieving state-of-the-art results among all evaluated methods. The source code and experimental dataset are available at \url{https://github.com/CastleChen339/DehazeXL}.
\end{abstract}    
 \section{Introduction}
\label{sec:intro}
\vspace{-4pt}
Image dehazing is a critical operation in various applications, including surveillance~\cite{suresh2024enhanced, jackson2024hazy}, autonomous navigation~\cite{kim2024modified, saravanarajan2023improving}, and remote sensing~\cite{xu2024mrf}. Haze significantly degrades image quality by obscuring details and distorting color representation, which impairs the performance of subsequent visual tasks such as object detection~\cite{sharma2023review, rani2024traffic} and tracking~\cite{sharma2023review}. In order to address this issue, researchers have developed a multitude of approaches that leverage Convolutional Neural Networks (CNNs)~\cite{zheng2023curricular, lu2024mixdehazenet, cui2024revitalizing, Zhang_2024_CVPR}, Generative Adversarial Networks (GANs)~\cite{zhang2022research, tassew2024dc, zhang2024gan}, Transformers~\cite{guo2022image, song2023vision, qiu2023mb}, and Diffusion models~\cite{xie2024frequency, cheng2024dehazediff} to tackle the haze removal problem. These methods have demonstrated exceptional performance in various fields, successfully restoring clarity and improving visual fidelity.

\begin{figure}[t]
  \centering
   \includegraphics[width=\linewidth]{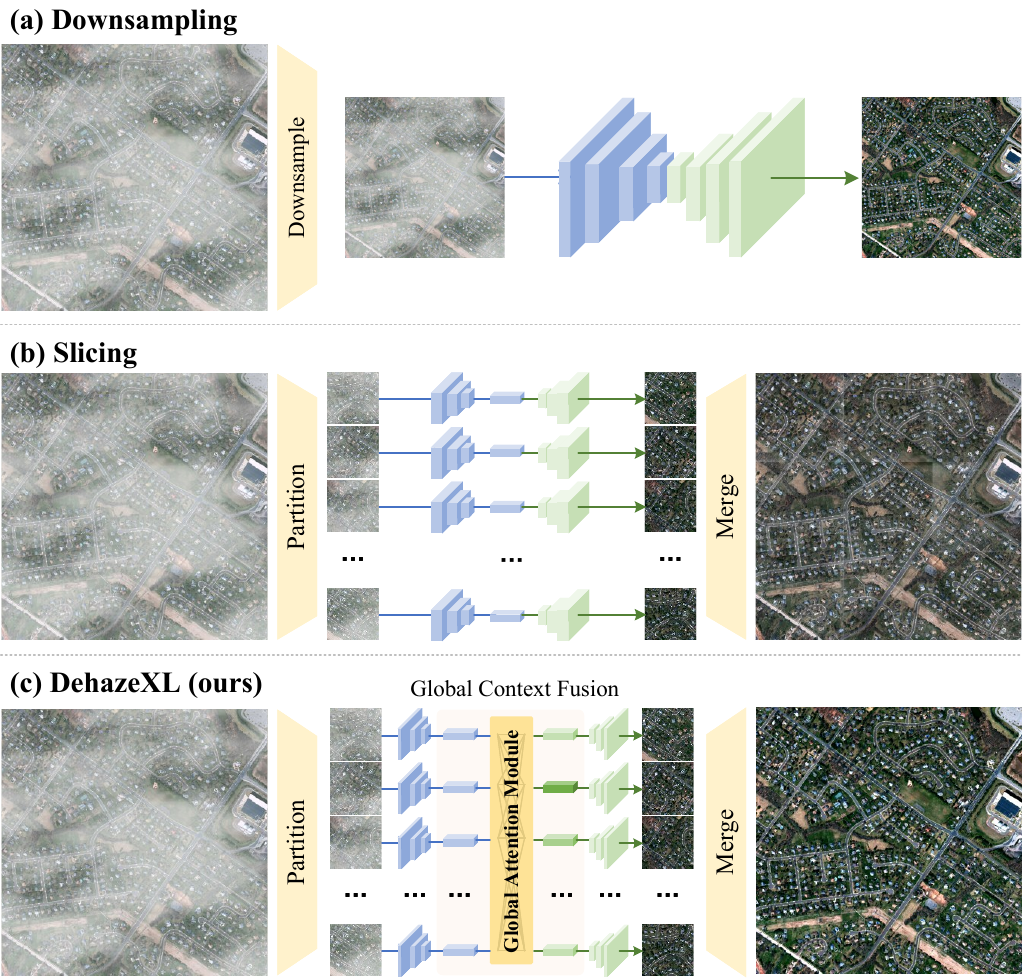}
   \caption{Comparison between different methods for handling large images in haze removal tasks. (a) Downsampling approach, which reduces the image size but loses critical high-frequency details. (b) Image slicing technique, which processes larger inputs but compromises global contextual information and object coherence. (c) The proposed method, which aims to effectively balance global context and local feature extraction to enhance haze removal performance in high-resolution images.}
   \label{fig:first}
\end{figure}

With advancements in image sensor technology, both the resolution and scale of captured images are steadily increasing. However, most existing dehazing methods have been developed and tested on relatively small images, typically ranging from 256 $\times$ 256 to 512 $\times$ 512 pixels. Constrained by GPU memory, these methods often employ compromises when processing large inputs, resorting to strategies such as slicing and downsampling~\cite{jin2021learning, shan2021uhrsnet, zheng2021ultra, gupta2024xt}. Although image slicing allows the processing of large inputs, it disrupts global contextual information, potentially leading to a loss in object coherence and spatial relationships. On the other hand, downsampling preserves global structure but sacrifices critical high-frequency details that are vital for downstream tasks such as object detection. These limitations highlight the need for innovative solutions that can efficiently balance global context and local details in the haze removal domain, particularly for high-resolution imagery.

\begin{figure}[t]
  \centering
   \includegraphics[width=\linewidth]{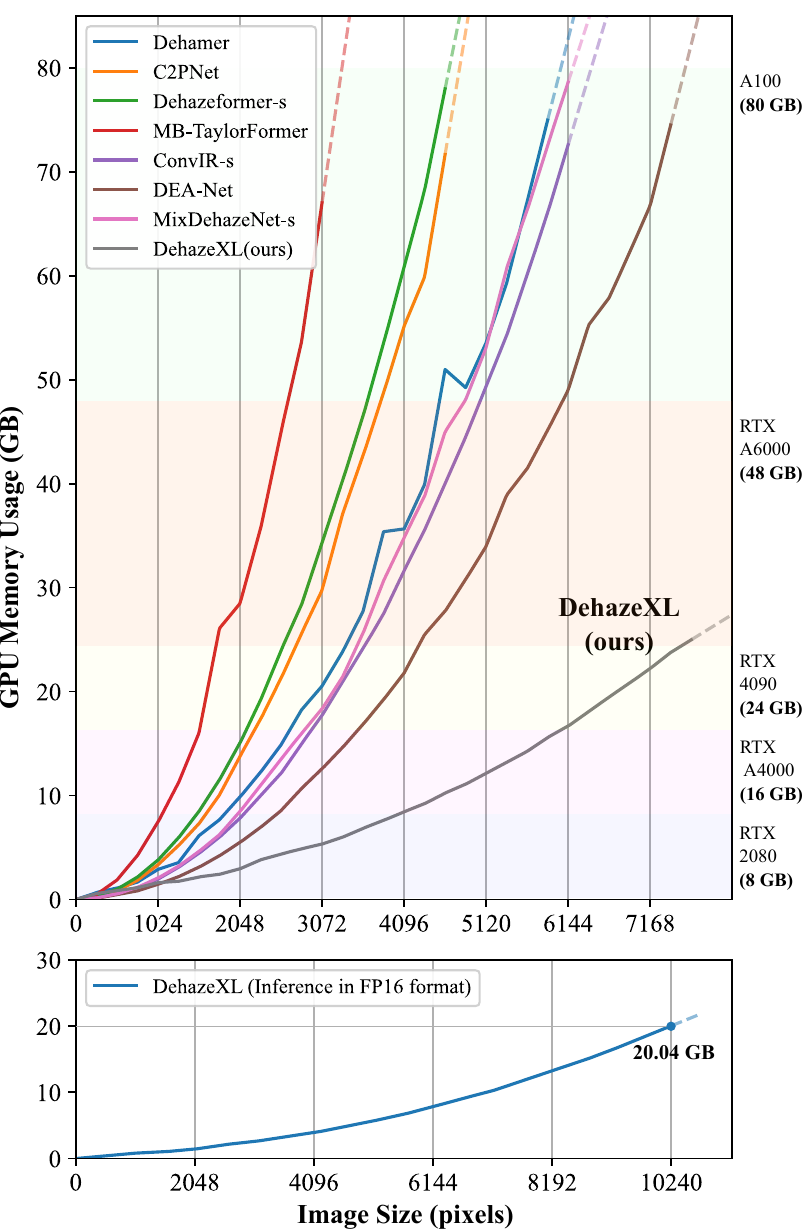}
   \setlength{\abovecaptionskip}{-8pt}
   \setlength{\belowcaptionskip}{-12pt}
   \caption{Comparison of GPU memory usage across various models. DehazeXL demonstrates a reduction in memory usage by approximately 65\%-80\% when processing large images compared to other methods. Notably, when employing FP16 format for inference, DehazeXL can process 10,240 $\times$ 10,240 pixel images with only 21 GB of memory.}
   \label{fig:mem}
\end{figure}

In this paper, we propose DehazeXL, an end-to-end haze removal method that effectively integrates global information interaction with local details extraction. As shown in Figure \ref{fig:mem}, DehazeXL is capable of directly inferring large images without incurring quadratic increases in GPU memory usage. Specifically, the input image is partitioned into equal-sized patches, each encoded into a feature vector by a shared encoder. These feature vectors serve as tokens for the global attention module, facilitating integration of broader contextual information. The globally enhanced features are then passed through a decoder, progressively upsampled to the original patch size, and finally merged to generate the output image.

The key features of DehazeXL are characterized by three aspects: 
\textbf{1) Decoupled Input Dimensions.} By partitioning images into fixed-size patches, DehazeXL decouples the encoder-decoder input dimensions from the image size. This approach enables efficient batch processing of image patches while conserving GPU memory, mitigating the risk of memory overflow. Moreover, maintaining a consistent patch size standardizes inputs for both the encoder and decoder, which enhances training stability and convergence.
\textbf{2) Enhanced Local Feature Representation.} A customized global attention module enriches each local feature vector with essential global context, which includes haze distribution, color consistency in clear regions, and brightness levels. This information is vital for accurate scene reconstruction. Without the global information, local feature vectors may lack spatial coherence, potentially leading to artifacts or inconsistencies in the output. 
\textbf{3) Efficient Global Attention Mechanisms.} Drawing inspiration from long-context attention mechanisms in large language models, we incorporate locality-sensitive hashing and low-rank decomposition into our global attention module. This design reduces the memory usage and computational demands when processing long contexts, thereby improving the model's ability to capture extensive contextual dependencies across ultra-high-resolution images.


Compared to existing methods, the most significant advancement of DehazeXL lies in its efficient global modeling capability for large inputs. To investigate the impact of global information utilization efficiency on dehazing performance, we develop a visual attribution method specifically tailored for haze removal tasks. By analyzing the contribution of each region, we can gain insights into which features are most influential in haze removal, thereby enhancing our understanding of the underlying processes involved. This approach not only facilitates the optimization of model performance but also provides a framework for interpreting results, which is crucial for advancing research in the field.


Additionally, we unexpectedly discovered a notable scarcity of ultra-high-resolution datasets designed for haze removal through extensive literature review. The existing datasets, such as 4KID~\cite{zheng2021ultrahigh}, are limited to a maximum resolution of 3840 $\times$ 2160 pixels. To fill this gap, we construct a haze removal dataset (\emph{8KDehaze}) using aerial images. Unlike existing haze removal datasets, all images in \emph{8KDehaze} have a resolution of 8192 $\times$ 8192 pixels, providing a unique resource for training and evaluating dehazing algorithms on ultra-high-resolution data. 



In summary, our key contributions are as follows:
\begin{itemize}
    \item We propose DehazeXL, an end-to-end haze removal method that seamlessly integrates global information interaction with local feature extraction. This approach allows for efficient processing of large images without significant increases in GPU memory usage.
    \item To evaluate the efficiency of global context utilization in haze removal performance, we design a visual attribution method called Dehazing Attribution Map (DAM). This method enables the identification and quantification of how specific regions or features contribute to model performance, supporting optimization and interpretability.
    \item We constuct an ultra-high-resolution haze removal dataset (\emph{8KDehaze}), which comprises images with a resolution of 8192 $\times$ 8192 pixels sourced from aerial imagery. This dataset addresses the scarcity of high-resolution resources in haze removal research and includes a diverse range of haze distributions and terrains, facilitating rigorous evaluation and future advancement of dehazing algorithms. 
\end{itemize}

 \vspace{-4pt}
\section{Related Work}
\label{sec:related}
\vspace{-4pt}
\noindent{\bf Single Image Dehazing.} Single image dehazing has progressed significantly over the past few decades. Traditional methods predominantly relied on atmospheric scattering models, utilizing handcrafted priors such as the Dark Channel Prior~\cite{he2010single, lee2016review} and Color Attenuation Prior~\cite{zhu2015fast, zhu2014single}. However, these methods often struggled in complex scenes due to oversimplified assumptions about scene structure and atmospheric conditions. The advent of large-scale hazy image datasets has catalyzed the rapid development of data-driven methods. Researchers have increasingly turned to deep learning models~\cite{qin2020ffa, dong2020multi, zheng2021ultra} to overcome the limitations of traditional techniques. Recent methods often incorporate attention mechanisms~\cite{li2023attention, tong2024haze}, Multi-scale feature fusion mechanisms~\cite{liu2023local, yang2023mstfdn}, and physically grounded models~\cite{zhou2023physical, he2023remote} to improve dehazing performance. The integration of deep learning not only enhances feature extraction capabilities but also facilitates the modeling of complex atmospheric phenomena. This transition to data-driven methodologies marks a great advancement in the field, enabling more accurate dehazing results. However, most deep learning-based dehazing methods struggle to infer high-resolution images due to GPU memory constraints, limiting their practical use in real-world applications. 

\noindent{\bf Large Image Inference.} With advancements in imaging sensor technologies, high-resolution image modeling and inference have emerged as key challenges in computer vision. Techniques for addressing large images typically fall into two categories: multi-scale hierarchical (or cascading) methods and sliding window strategies. R-CNN~\cite{girshick2014rich} and CNN cascades~\cite{ gadermayr2019cnn} demonstrated the effectiveness of cascading networks for large images, though at the cost of speed. Recently, Gupta et al.~\cite{xTLargeImageModeling} designed a visual backbone network for high-level vision tasks involving large images. They sliced the input images to extract local features and then employed a self-attention mechanism to derive global information from these local features. This approach achieved impressive performance in image classification, object detection, and segmentation tasks. In the domain of haze removal, Zheng et al.~\cite{zheng2021ultrahigh} proposed a model capable of processing 4K images on a single GPU by combining three CNNs for feature extraction, guidance map learning, and feature fusion. Conversely, sliding window methods are widely used in various visual tasks \cite{yan2025mitigating}, where large images are divided into smaller patches to enable localized processing. However, both approaches have inherent limitations. Multi-scale hierarchical methods suffer from memory usage that scales quadratically with input size, posing serious computational challenges. Sliding window techniques can disrupt spatial coherence in tasks like dehazing, leading to block artifacts at the edges of the windows. Balancing computational efficiency with contextual integrity remains an open research challenge.

\begin{figure*}[t]
  \centering
   \includegraphics[width=\linewidth]{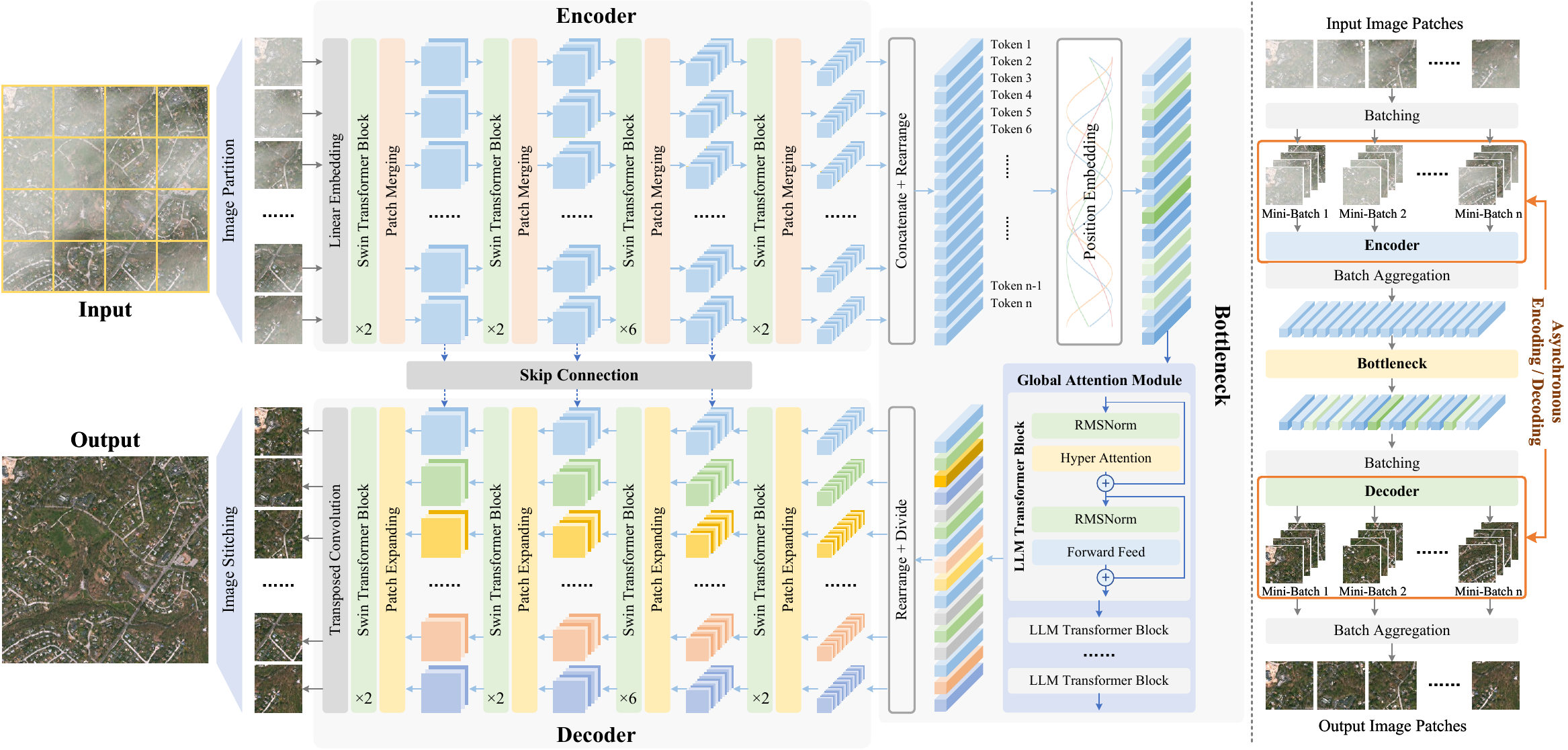}
   \setlength{\abovecaptionskip}{-8pt}
   \setlength{\belowcaptionskip}{-8pt}
   \caption{Overall architecture of the proposed model. It begins by partitioning the hazy image into uniform-sized patches, which are then encoded into tokens by the Encoder. The Bottleneck injects global information into each token, enhancing the contextual representation. Subsequently, the Decoder reconstructs the tokens back into image patches, forming the final output image. Notably, to minimize memory consumption, both the Encoder and Decoder employ an asynchronous processing strategy, handling the input in multiple mini-batches sequentially rather than simultaneously. This design optimizes memory efficiency while ensuring effective haze removal. }
   \label{fig:modelarch}
\end{figure*}

\noindent{\bf Visual Interpretation of Networks} As deep neural networks become increasingly prevalent in computer vision, there has been growing interest in understanding the factors that influence their outputs. This process, known as attribution analysis, aims to provide insight into which features contribute most significantly to the network's decisions. Over recent years, numerous attribution methods~\cite{nielsen2022robust, achtibat2023attribution, zheng2024attribution, gevaert2024evaluating} have been developed to produce interpretable and intuitive visual explanations. Some works~\cite{zhao2024gradient, yang2023idgi} focus on analyzing the internal parameters of the network, tracing how information flows through layers and nodes to attribute predictions. However, this becomes challenging for highly complex models due to the intricate nature of their architectures. To address this, other methods~\cite{fel2023don, arumugam2023interpreting} treat the network as a black box, perturbing key features of the input to assess their impact on the output. This perturbation-based approach evaluates the sensitivity of the model to specific input regions or features, offering a more flexible means of interpretation without requiring detailed knowledge of the model’s inner workings. In addition, there are also works on improving model interpretability, such as Local Attribution Map~\cite{gu2021interpreting} and LIME~\cite{yang2023investigating, nagahisarchoghaei2023generative}, which provides localized explanations by approximating the complex model's predictions with simpler, interpretable models in the vicinity of specific instances.

 \vspace{-4pt}
\section{Methodology}
\vspace{-4pt}
The key contribution of our work is to design an end-to-end haze removal model for large images. The architecture and details of the proposed DehazeXL are presented in Section \ref{para:model}. In addition, we develop a visual attribution method for dehazing tasks called DAM. Section \ref{para:dam} presents the principles of this method.

\subsection{Architecture of DehazeXL}
\vspace{-3pt}
\label{para:model}
As shown in Figure \ref{fig:modelarch}, the framework of DehazeXL consists of three primary components: the Encoder, Bottleneck, and Decoder. Initially, the hazy input image is divided into several fixed-size patches. These patches are then input into the Encoder for tokenization. The Bottleneck is designed to inject global information into each token, thereby enhancing their contextual representation. Finally, the Decoder reconstructs the processed tokens into patches, resulting in the final dehazed image.

\noindent{\bf Encoder.} 
The Encoder can be any visual model backbone capable of extracting local features from each image patch. In our experiments, we employed the Swin Transformer V2~\cite{liu2022swin2} as the Encoder. This choice leverages the Swin Transformer’s ability to capture hierarchical features and its efficient handling of long-range dependencies, which is particularly advantageous for processing complex hazy images. Since the Encoder focuses solely on local features, we adopt a strategy of dividing the patches into multiple mini-batches for sequential input to the Encoder, rather than processing all patches simultaneously. While this design may slow down the encoding speed, it effectively decouples the memory usage of the Encoder from the size of the input image, significantly reducing memory consumption and enabling the processing of large-scale images.

\noindent{\bf Bottleneck.} 
Within the Encoder, all patches are encoded into smaller feature maps, referred to as tokens. These tokens are then input into the Bottleneck. We constructed an efficient Transformer block to extract global information and inject it into all tokens. We utilized RMSNorm~\cite{zhang2019root} as the normalization layer to save computational time. Additionally, inspired by large language models~\cite{touvron2023llama}, we implemented Hyper Attention~\cite{han2023hyperattention}, which was confirmed to be effective in natural language processing. This approach aims to enhance inference speed while minimizing memory usage, particularly for long-context inputs. Consequently, all tokens can "see" each other, facilitating the learning of global information such as haze distributions, color characteristics, and brightness.

\noindent{\bf Decoder.} 
The Decoder's function is to reconstruct the tokens into clear, haze-free patches. Similar to the Encoder, we utilized the Swin Transformer V2~\cite{liu2022swin2} as the backbone, substituting the Patch Merging layer with a Patch Expanding layer that employs transposed convolution to iteratively upscale and merge feature maps. Through skip connections, we concatenate the outputs from each layer of the Encoder with the corresponding feature maps in the Decoder, thus enhancing the flow of information and gradients. Consistent with our approach in the Encoder, we adopt a "divide and conquer" strategy in the Decoder, sequentially processing all tokens instead of concurrently. This strategy allows us to achieve significantly lower memory usage at the cost of slightly increased processing time.

\subsection{Dehazing Attribution Map}
\label{para:dam}
\vspace{-3pt}
Inspired by the Integrated Gradients (IG) method~\cite{sundararajan2017axiomatic} and the Local Attribution Map~\cite{gu2021interpreting}, we propose the Dehazing Attribution Map to enhance the interpretability of our model. Let $F: \mathbb{R}^{h\times w} \to \mathbb{R}^{h\times w} $ represent a dehazing network. To quantify the dehazing effect, we utilize a pixel intensity detector, given the significant differences in pixel intensities between hazy and clear images. Specifically, for an input hazy image $I \in \mathbb{R}^{h\times w}$, we define the detector as $D_{xy}(I)= {\textstyle \sum_{i\in [x,x+l],j\in [y,y+l]}^{}} I_{ij}$, where the subscripts $i$ and $j$ denote the spatial coordinates. For clarity, we will omit the subscripts in the subsequent discussion.
To conduct attribution analysis for dehazing network, we require a baseline input image ${I}'$ which satisfies that $F({I}')$ absent certain features present in $F(I)$. The attribution map $D(F(I))$ is obtained by computing the path-integrated gradient along a continuous trajectory transitioning from ${I}'$ to $I$. This smooth path function is denoted as $\gamma (\alpha ) : [0, 1] \to \mathbb{R}^{h\times w}$, with $\gamma (0): {I}'$ and  $\gamma (1): I$. Therefore, the $i$-th dimension of the attribution map can be expressed as follows:

\begin{equation}
  DAM_{F,D}(\gamma)_{i} = \int_{0}^{1}\frac{\partial D(F(\gamma(\alpha)))}{\partial\gamma (\alpha)_{i}} \times  \frac{\partial\gamma(\alpha )_{i}}{\partial\alpha }d\alpha
  \label{eq:dam}
  \abovedisplayshortskip=0pt
  \abovedisplayskip=0pt
\end{equation}

\begin{figure}[t]
  \centering
   \includegraphics[width=\linewidth]{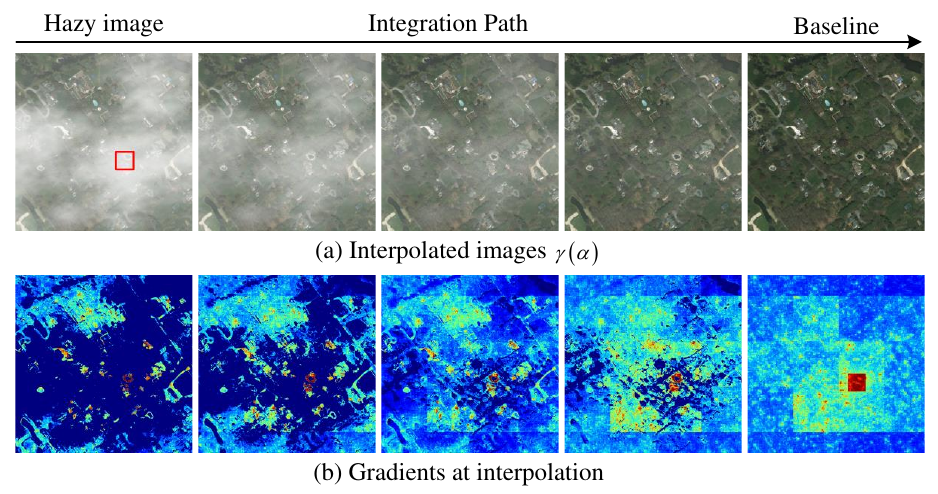}
   \setlength{\abovecaptionskip}{-4pt}
   \setlength{\belowcaptionskip}{-8pt}
   \caption{Illustration of the baseline image and the path function. The region enclosed by the red box indicates the attribution area.}
   \label{fig:grad}
\end{figure}

As highlighted in \cite{sturmfels2020visualizing}, the effectiveness of model attribution depends on the choice of an appropriate baseline. For instance, in image classification tasks, a pure black image serves as a suitable baseline since the model is unable to classify it~\cite{sundararajan2017axiomatic}. In this work, we meticulously design baseline inputs specifically tailored for the dehazing network. As stated above, a baseline input must lack certain key features, which are typically determined by the characteristics of the task. In the context of dehazing, clear regions of an image are easy to reconstruct. In contrast, reconstructing hazy regions, particularly those with thick haze, poses substantial challenges. Effectively reconstructing these hazy areas is crucial for achieving superior dehazing results. Therefore, as shown in Figure \ref{fig:grad}, we utilize the clear image as the baseline input and adopt a linear interpolation function as the path function. In practice, we compute the gradients at uniformly sampled points along the defined path, then approximating the integral as described in Eq (\ref{eq:app}):

\begin{equation}
  \tilde{DAM}_{F,D}(\gamma)_{i} = \sum_{k= 1}^{m}\frac{\partial D(F(\gamma (\frac{k}{m}) ))}{\partial \gamma(\frac{k}{m})_{i} } \cdot \frac{(\Delta \gamma_{k,m})_{i}}{m}
  \label{eq:app}
\abovedisplayshortskip=0pt
\abovedisplayskip=0pt
\end{equation}

\noindent where $m$ denotes the number of steps used for the integral approximation, and $\Delta \gamma_{k,m} = \gamma(\frac{k}{m} )-\gamma(\frac{k+1}{m})$ .Empirically, we find that a step count of 100 is sufficient to approximate the integral effectively.

 \section{Experiments and Analysis}

\subsection{Dataset}
\vspace{-3pt}
To train and evaluate the proposed network and the comparative methods, we constructed an ultra-high-resolution haze removal dataset (\emph{8KDehaze}), containing 10,000 images at a resolution of 8192 $\times$ 8192 pixels. The clear images in \emph{8KDehaze} were sourced from publicly available aerial imagery provided by the United States Geological Survey, while the hazy counterparts were generated using the atmospheric scattering model~\cite{he2010single} and the approach proposed by Czerkawski et al.\cite{SatelliteCloudGenerator}.
To the best of our knowledge, \emph{8KDehaze} is the first ultra-high-resolution dataset in the field of image dehazing. In addition, to further validate the effectiveness of the proposed method, we conduct extensive training and testing on the synthesis dataset \emph{4KID}~\cite{zheng2021ultra} and the real-world dataset \emph{O-HAZE}~\cite{ancuti2018ohaze}. Table \ref{tab:data} shows the details of datasets used in the experiments.
\vspace{-3mm} 

\begin{table}[ht]
\setlength{\abovecaptionskip}{3pt}
\setlength{\belowcaptionskip}{0pt}
\centering
\caption{Overview of datasets used in the experiments.}
\setlength{\tabcolsep}{1mm}
\fontsize{8}{12}\selectfont{\begin{tabular}{c||ccc}
\hline
\textbf{Dataset}    & \textbf{4KID}       & \textbf{O-HAZY}                                                                      & \textbf{8KDehaze}                                                                                  \\ \hline
\textbf{Quantity}   & 15606               & 45                                                                                   & 10000                                                                                              \\ \hline
\textbf{Image Size} & 3840 $\times$ 2160  & \begin{tabular}[c]{@{}c@{}}1286 $\times$ 947 \\ to\\ 5436 $\times$ 3612\end{tabular} & 8192 $\times$ 8192                                                                                 \\ \hline
\textbf{Source}     & Video Frames & Commercial Camera                                                                    & Aerial Images                                                                                      \\ \hline
\textbf{Content}    & Urban Streets       & Parks, Suburban                                                                      & \begin{tabular}[c]{@{}c@{}}Urban, Farmland,\\ Mountains, Desert,\\ Coastlines, Rivers\end{tabular} \\ \hline
\end{tabular}
\label{tab:data}}
\end{table}

\begin{figure*}[t]
  \centering
   \includegraphics[width=\linewidth]{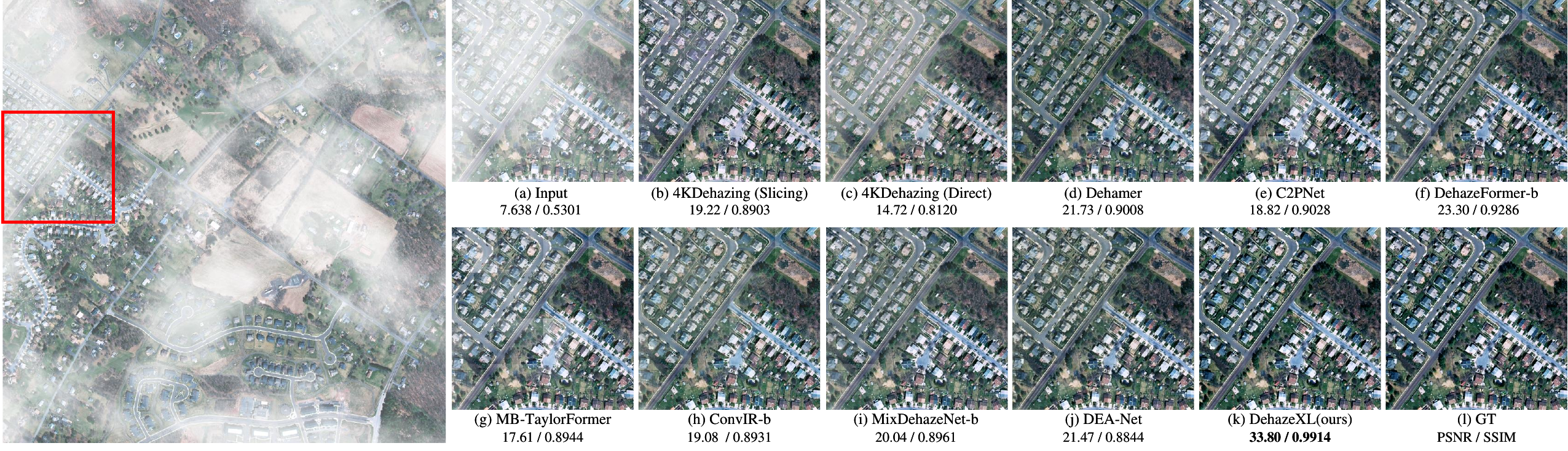}
   \setlength{\abovecaptionskip}{-5pt}
   \setlength{\belowcaptionskip}{0pt}
   \caption{Dehazed results on the \emph{8KDehaze} dataset. The patches for comparison are marked with red boxes in the original images. PSNR / SSIM is calculated based on the patches to better reflect the performance difference. The proposed DehazeXL can directly infer images with a resolution of 8192 $\times$ 8192 without the need for slicing inference. Compared to other methods, the proposed method effectively eliminates segmentation artifacts and achieves superior visual quality.}
   \label{fig:res1}
\end{figure*}

\begin{figure*}[t]
  \centering
   \includegraphics[width=\linewidth]{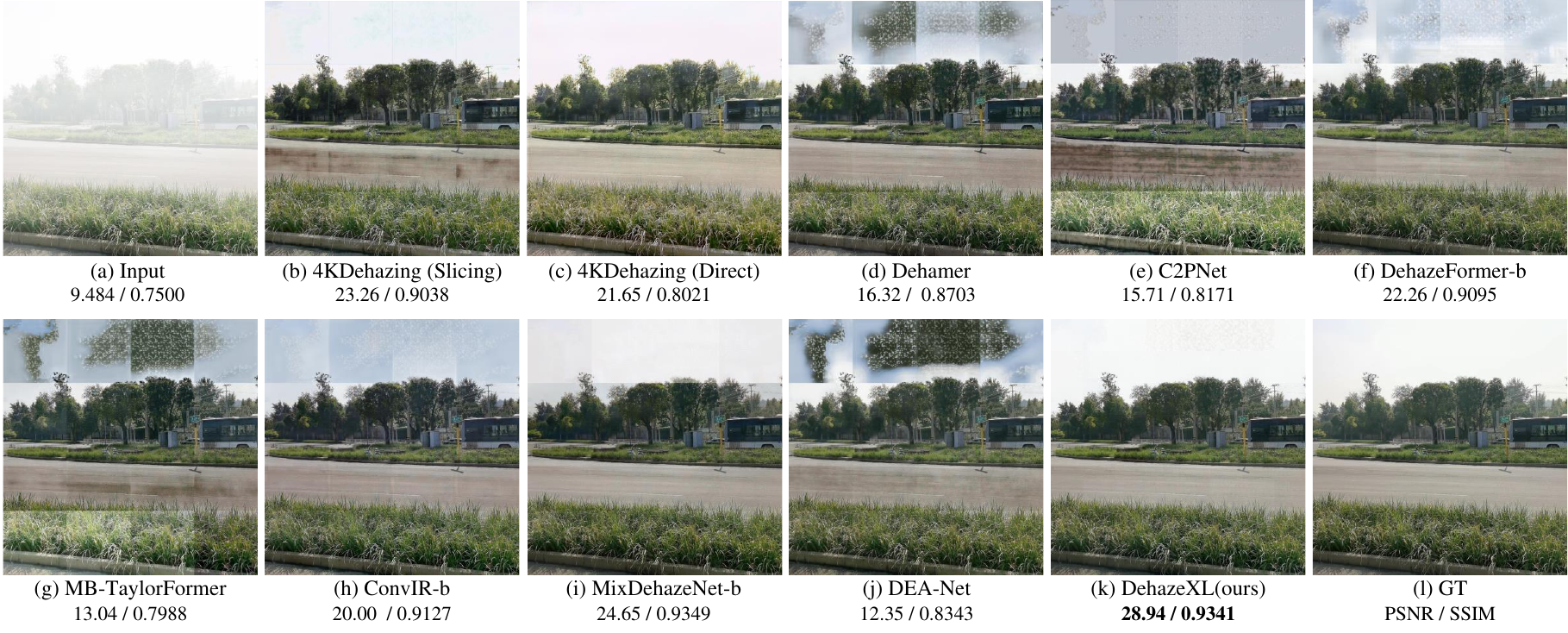}
   \setlength{\abovecaptionskip}{-5pt}
   \setlength{\belowcaptionskip}{-15pt}
   \caption{Dehazed results on the \emph{4KID}~\cite{zheng2021ultra} dataset. The proposed DehazeXL can effectively utilize global information to guide image restoration in different regions, enhancing the global consistency of the output results.}
   \label{fig:res4kid}
\end{figure*}

\begin{figure*}[t]
  \centering
   \includegraphics[width=\linewidth]{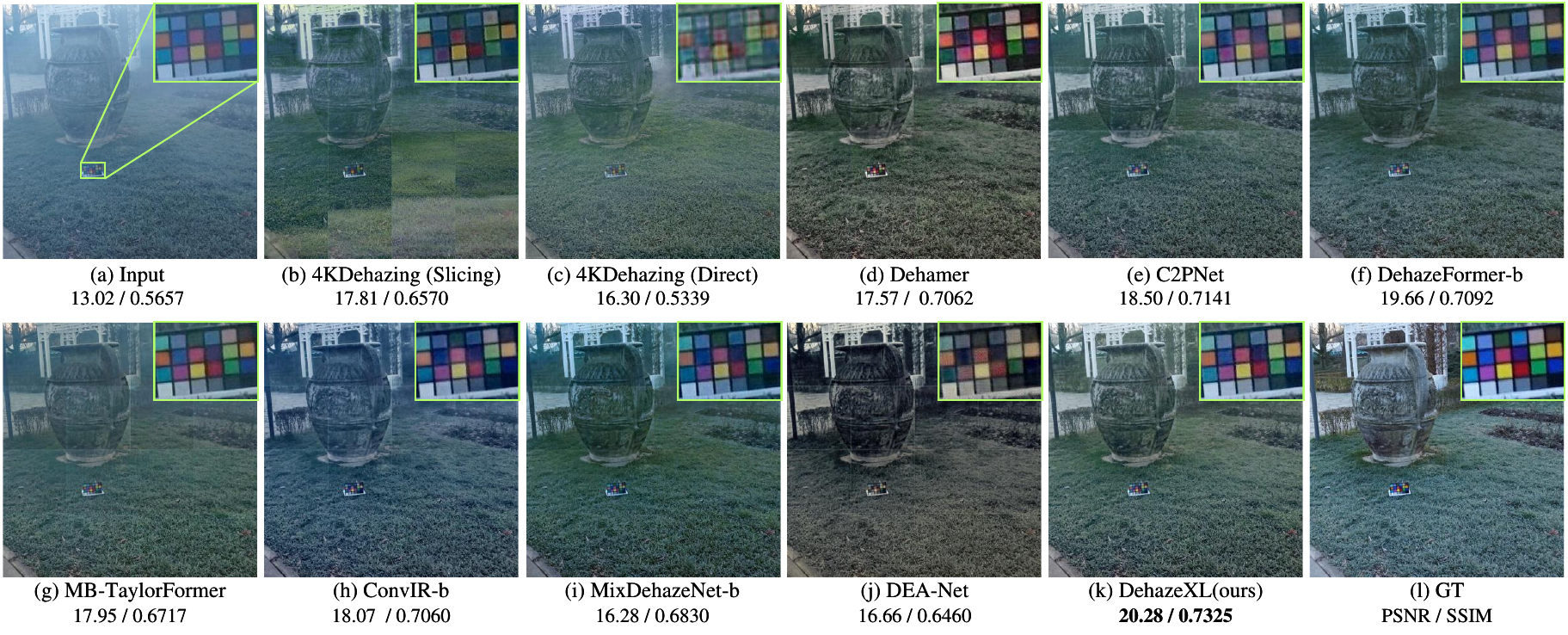}
   \setlength{\abovecaptionskip}{-5pt}
   \setlength{\belowcaptionskip}{-5pt}
   \caption{Dehazed results on the \emph{O-HAZE}~\cite{ancuti2018ohaze} dataset. The proposed DehazeXL demonstrates higher color fidelity and restores more details compared with other state-of-the-art methods.}
   \label{fig:resohaze}
\end{figure*}

\begin{table*}[ht]
\centering
\caption{Quantitative evaluations on the 8KDehaze dataset, the 4KID dataset~\cite{zheng2021ultra}, and the O-HAZE dataset~\cite{ancuti2018ohaze} in terms of PSNR, SSIM, and average infer time.}
\setlength{\tabcolsep}{2.1mm}
\fontsize{9}{13}\selectfont{\begin{tabular}{cc|ccc|ccc|ccc}
\hline
\multicolumn{1}{c|}{\multirow{2}{*}{\textbf{Method}}} & \multirow{2}{*}{\textbf{Venue\&Year}} & \multicolumn{3}{c|}{\textbf{8KDehaze}}                                                        & \multicolumn{3}{c|}{\textbf{4KID}~\cite{zheng2021ultra}}                                                            & \multicolumn{3}{c}{\textbf{O-HAZY}~\cite{ancuti2018ohaze}}                                                           \\ \cline{3-11} 
\multicolumn{1}{c|}{}                                 &                                       & \multicolumn{1}{c|}{\textbf{PSNR}}  & \multicolumn{1}{c|}{\textbf{SSIM}}   & \textbf{Time(s)} & \multicolumn{1}{c|}{\textbf{PSNR}}  & \multicolumn{1}{c|}{\textbf{SSIM}}   & \textbf{Time(s)} & \multicolumn{1}{c|}{\textbf{PSNR}}  & \multicolumn{1}{c|}{\textbf{SSIM}}   & \textbf{Time(s)} \\ \hline
\multicolumn{1}{c|}{4KDehazing (Slicing)}             & CVPR2021                              & \multicolumn{1}{c|}{25.81}          & \multicolumn{1}{c|}{0.9569}          & 6.682            & \multicolumn{1}{c|}{19.97}          & \multicolumn{1}{c|}{0.8624}          & 1.04             & \multicolumn{1}{c|}{18.73}          & \multicolumn{1}{c|}{0.6726}          & 1.31             \\
\multicolumn{1}{c|}{4KDehazing (Direct)}              & CVPR2021                              & \multicolumn{1}{c|}{20.41}          & \multicolumn{1}{c|}{0.8664}          & 1.350            & \multicolumn{1}{c|}{18.68}          & \multicolumn{1}{c|}{0.7424}          & 0.19             & \multicolumn{1}{c|}{19.3}           & \multicolumn{1}{c|}{0.6426}          & 0.27             \\
\multicolumn{1}{c|}{Dehamer}                          & CVPR2022                              & \multicolumn{1}{c|}{25.92}          & \multicolumn{1}{c|}{0.9373}          & 6.614            & \multicolumn{1}{c|}{21.24}          & \multicolumn{1}{c|}{0.8795}          & 1.03             & \multicolumn{1}{c|}{19.59}          & \multicolumn{1}{c|}{0.7134}          & 1.30             \\
\multicolumn{1}{c|}{C2PNet}                           & CVPR2023                              & \multicolumn{1}{c|}{26.17}          & \multicolumn{1}{c|}{0.9669}          & 43.269           & \multicolumn{1}{c|}{18.14}          & \multicolumn{1}{c|}{0.8299}          & 6.76             & \multicolumn{1}{c|}{20.29}          & \multicolumn{1}{c|}{0.7113}          & 8.51             \\
\multicolumn{1}{c|}{DehazeFormer-s}                   & TIP2023                               & \multicolumn{1}{c|}{26.68}          & \multicolumn{1}{c|}{0.9729}          & 7.469            & \multicolumn{1}{c|}{20.83}          & \multicolumn{1}{c|}{0.8763}          & 1.17             & \multicolumn{1}{c|}{19.86}          & \multicolumn{1}{c|}{0.7116}          & 1.47             \\
\multicolumn{1}{c|}{DehazeFormer-b}                   & TIP2023                               & \multicolumn{1}{c|}{26.83}          & \multicolumn{1}{c|}{0.9657}          & 15.013           & \multicolumn{1}{c|}{21.25}          & \multicolumn{1}{c|}{0.8843}          & 2.35             & \multicolumn{1}{c|}{20.22}          & \multicolumn{1}{c|}{0.7173}          & 2.95             \\
\multicolumn{1}{c|}{MB-TaylorFormer}                  & ICCV2023                              & \multicolumn{1}{c|}{26.41}          & \multicolumn{1}{c|}{0.9668}          & 120.540          & \multicolumn{1}{c|}{18.63}          & \multicolumn{1}{c|}{0.8497}          & 18.83            & \multicolumn{1}{c|}{19.57}          & \multicolumn{1}{c|}{0.7104}          & 23.71            \\
\multicolumn{1}{c|}{ConvIR-s}                         & TPAMI2024                             & \multicolumn{1}{c|}{25.11}          & \multicolumn{1}{c|}{0.9599}          & 6.661            & \multicolumn{1}{c|}{20.66}          & \multicolumn{1}{c|}{0.8696}          & 1.04             & \multicolumn{1}{c|}{18.83}          & \multicolumn{1}{c|}{0.7095}          & 1.31             \\
\multicolumn{1}{c|}{ConvIR-b}                         & TPAMI2024                             & \multicolumn{1}{c|}{26.93}          & \multicolumn{1}{c|}{0.9775}          & 8.709            & \multicolumn{1}{c|}{21.92}          & \multicolumn{1}{c|}{0.888}           & 1.36             & \multicolumn{1}{c|}{19.61}          & \multicolumn{1}{c|}{0.7199}          & 1.71             \\
\multicolumn{1}{c|}{MixDehazeNet-s}                   & IJCNN2024                             & \multicolumn{1}{c|}{20.99}          & \multicolumn{1}{c|}{0.8934}          & 6.563            & \multicolumn{1}{c|}{21.25}          & \multicolumn{1}{c|}{0.8817}          & 1.03             & \multicolumn{1}{c|}{19.09}          & \multicolumn{1}{c|}{0.7165}          & 1.29             \\
\multicolumn{1}{c|}{MixDehazeNet-b}                   & IJCNN2024                             & \multicolumn{1}{c|}{23.16}          & \multicolumn{1}{c|}{0.9284}          & 13.154           & \multicolumn{1}{c|}{23.22}          & \multicolumn{1}{c|}{0.9063}          & 2.06             & \multicolumn{1}{c|}{20.67}          & \multicolumn{1}{c|}{0.7293}          & 2.59             \\
\multicolumn{1}{c|}{DEA-Net}                          & TIP2024                               & \multicolumn{1}{c|}{25.89}          & \multicolumn{1}{c|}{0.9329}          & 7.402            & \multicolumn{1}{c|}{20.83}          & \multicolumn{1}{c|}{0.8834}          & 1.16             & \multicolumn{1}{c|}{20.01}          & \multicolumn{1}{c|}{0.6988}          & 1.46             \\ \hline
\multicolumn{2}{c|}{\textbf{DehazeXL}}                                                        & \multicolumn{1}{c|}{\textbf{32.35}} & \multicolumn{1}{c|}{\textbf{0.9863}} & \textbf{4.617}   & \multicolumn{1}{c|}{\textbf{26.62}} & \multicolumn{1}{c|}{\textbf{0.9073}} & \textbf{0.59}    & \multicolumn{1}{c|}{\textbf{21.49}} & \multicolumn{1}{c|}{\textbf{0.7348}} & \textbf{0.86}    \\ \hline
\end{tabular}
\vspace{-10pt} 
\label{tab:per}}
\end{table*}
\vspace{-4mm} 

\subsection{Implementation Details}
\vspace{-3pt}
The proposed model was implemented in PyTorch and trained on a single NVIDIA A100 GPU. Input images were randomly cropped to a resolution of 2048 $\times$ 2048 pixels, with a batch size of 2 during training. For comparative analysis, we selected a range of recently published state-of-the-art dehazing algorithms, including 4KDehazing~\cite{zheng2021ultra}, Dehamer~\cite{guo2022image}, C2PNet~\cite{zheng2023curricular}, Dehazeformer~\cite{song2023vision}, MB-TaylorFormer~\cite{qiu2023mb}, ConvIR~\cite{cui2024revitalizing}, Mixdehazenet~\cite{lu2024mixdehazenet}, and DEA-Net~\cite{chen2024dea}. Since these approaches could not be trained directly on images at the 2048 $\times$ 2048 resolution, input image pairs were randomly cropped into patches of size 512 $\times$ 512 pixels. The training batch size for these methods was maximized based on available GPU memory. All models were trained using the Adam optimizer~\cite{kingma2014adam} with an initial learning rate of 0.001. To facilitate effective training, a cosine annealing schedule was employed to gradually decay the learning rate throughout the training process. Each model was trained for a total of 500 epochs, utilizing the L1 loss function as the objective.

In the testing phase, most comparative methods, including Dehamer~\cite{guo2022image}, C2PNet~\cite{zheng2023curricular}, Dehazeformer~\cite{song2023vision}, MB-TaylorFormer~\cite{qiu2023mb}, ConvIR~\cite{cui2024revitalizing}, MixdehazeNet~\cite{lu2024mixdehazenet}, and DEA-Net~\cite{chen2024dea}, employed a slicing inference strategy due to their limitations in processing large images. Notably, 4KDehazing~\cite{zheng2021ultra} is the only comparative method that supports direct inference on large images. Thus, the results for 4KDehazing were obtained using both slicing and direct inference. The proposed DehazeXL directly inferred the input images without employing the slicing strategy.

\begin{figure*}[t]
  \centering
    \includegraphics[width=\linewidth]{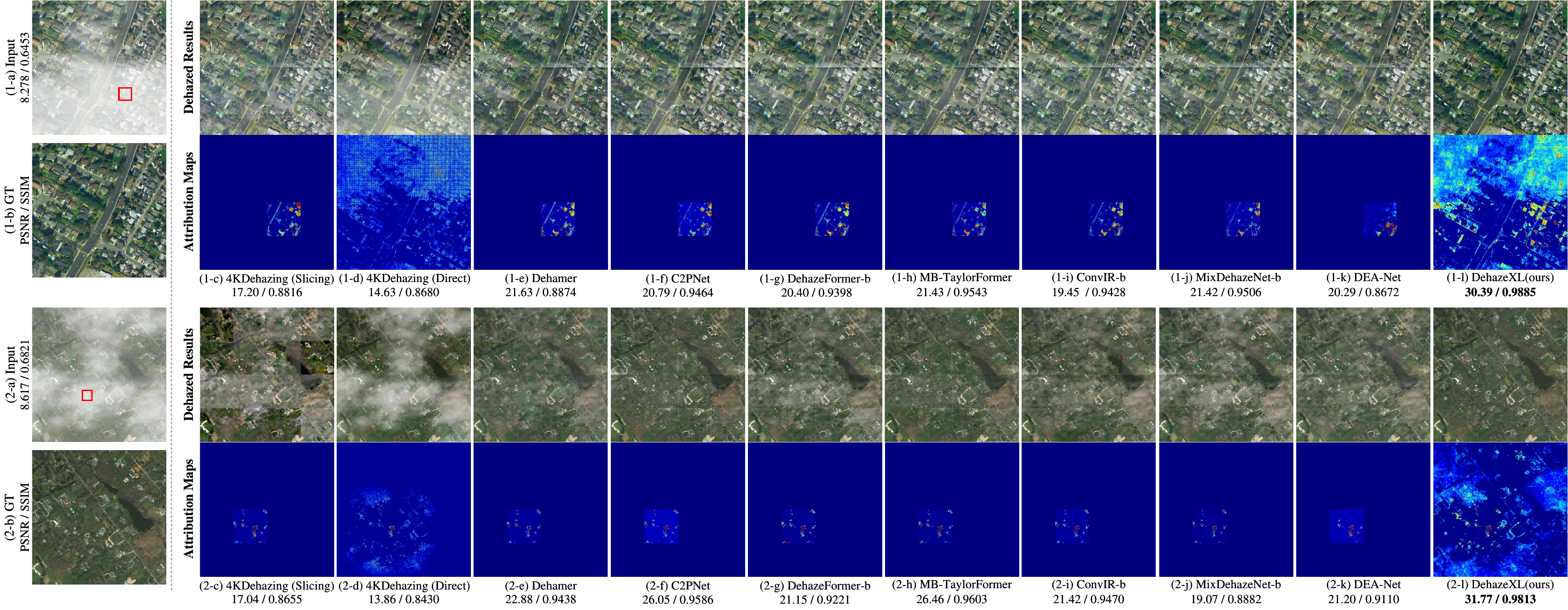}
    \setlength{\abovecaptionskip}{-5pt}
   \setlength{\belowcaptionskip}{-12pt}
   \caption{
   Comparison of the dehazed results and attribution maps of different methods. The red box on (1-a) and (2-a) indicate the regions of interest for attribution. The attribution maps highlight how each pixel influences the dehazing results in the specified region.
}
   \label{fig:lam}
\end{figure*}

\subsection{Evaluation and Results}
\vspace{-3pt}
\noindent\textbf{Qualitative Evaluation.}
Figure \ref{fig:res1} to \ref{fig:resohaze} present the testing results of the proposed method and comparative algorithms applied to samples from the \emph{8KDehaze}, \emph{4KID} and \emph{O-HAZE} datasets. As illustrated in Figure \ref{fig:res1}, methods that employ the slicing inference strategy exhibit noticeable block artifacts. While 4KDehazing can perform direct inference without the need for slicing, its dehazing performance significantly deteriorates when handling large images. In contrast, the proposed DehazeXL demonstrates superior dehazing capabilities. In Figure \ref{fig:res4kid}, all comparative methods exhibit varying degrees of failure, particularly in the sky regions. This is primarily due to the similarity in features between the sky and dense haze, which makes it challenging for slice-based inference methods to distinguish between sky regions and those obscured by haze. In contrast, DehazeXL effectively utilizes global information to differentiate the sky from hazy regions, thereby enhancing the global consistency of the output results. 
Figure \ref{fig:resohaze} further highlights the advantages of DehazeXL in terms of color restoration and overall coherence, demonstrating excellent generalization capability of the proposed method in real hazy scenes.
\noindent\textbf{Quantitative Evaluation.}
Table \ref{tab:per} summarizes the quantitative evaluation results of DehazeXL and the comparative methods on the \emph{8KDehaze}, \emph{4KID} and \emph{O-HAZE} datasets, using metrics such as PSNR, SSIM~\cite{wang2004image}, and average inference time. The proposed method achieves the highest scores for both PSNR and SSIM, indicating its superior dehazing effectiveness. Although 4KDehazing is faster with direct inference, it exhibits weaker performance on larger images and suffers from ghosting and color shifts. In contrast, DehazeXL achieves an excellent balance between dehazing performance and processing time, demonstrating its efficacy in practical applications.

\subsection{Ablation Study}
\vspace{-3pt}
We conducted ablation studies to evaluate the impact of different Backbone types and the depth of the Global Attention Module in the Bottleneck of the proposed DehazeXL. These experiments were performed on the 8KDehaze dataset, with the results presented in Table \ref{tab:data}. Our findings indicate that larger Backbone sizes and deeper Bottlenecks do indeed lead to improved performance; however, they also result in a significant increase in inference time. Considering the trade-off between inference time and model performance, we selected Swin-T and a depth of 2 as the default choices for the Backbone and Bottleneck, respectively.

\begin{table}[ht]
\setlength{\abovecaptionskip}{5pt}
\setlength{\belowcaptionskip}{-3pt}
\centering
\caption{Ablation study results for Backbone types and Bottleneck depth in DehazeXL on the 8KDehaze dataset.}
\setlength{\tabcolsep}{3mm}
\fontsize{8.5}{12}\selectfont{\begin{tabular}{c|c|ccc}
\hline
\multirow{2}{*}{\textbf{\begin{tabular}[c]{@{}c@{}}Backbone\\ Type\end{tabular}}} & \multirow{2}{*}{\textbf{\begin{tabular}[c]{@{}c@{}}Bottleneck\\ Depth\end{tabular}}} & \multicolumn{3}{c}{\textbf{Metrics}}                                                       \\ \cline{3-5} 
                                                                                  &                                                                                      & \multicolumn{1}{c|}{\textbf{PSNR}} & \multicolumn{1}{c|}{\textbf{SSIM}} & \textbf{Time(s)} \\ \hline
\multirow{3}{*}{Swin-T}                                                           & 1                                                                                    & \multicolumn{1}{c|}{31.61}         & \multicolumn{1}{c|}{0.9719}        & 4.511            \\
                                                                                  & 2                                                                                    & \multicolumn{1}{c|}{32.35}         & \multicolumn{1}{c|}{0.9863}        & 4.617            \\
                                                                                  & 4                                                                                    & \multicolumn{1}{c|}{32.40}         & \multicolumn{1}{c|}{0.9857}        & 4.810            \\ \hline
\multirow{3}{*}{Swin-S}                                                           & 1                                                                                    & \multicolumn{1}{c|}{31.89}         & \multicolumn{1}{c|}{0.9759}        & 5.880            \\
                                                                                  & 2                                                                                    & \multicolumn{1}{c|}{32.58}         & \multicolumn{1}{c|}{0.9871}        & 5.967            \\
                                                                                  & 4                                                                                    & \multicolumn{1}{c|}{32.76}         & \multicolumn{1}{c|}{0.9870}        & 6.102            \\ \hline
\multirow{3}{*}{Swin-B}                                                           & 1                                                                                    & \multicolumn{1}{c|}{32.10}         & \multicolumn{1}{c|}{0.9792}        & 9.066            \\
                                                                                  & 2                                                                                    & \multicolumn{1}{c|}{32.77}         & \multicolumn{1}{c|}{0.9879}        & 9.183            \\
                                                                                  & 4                                                                                    & \multicolumn{1}{c|}{33.06}         & \multicolumn{1}{c|}{0.9894}        & 9.526            \\ \hline
\multirow{3}{*}{Swin-L}                                                           & 1                                                                                    & \multicolumn{1}{c|}{32.93}         & \multicolumn{1}{c|}{0.9877}        & 17.37            \\
                                                                                  & 2                                                                                    & \multicolumn{1}{c|}{32.98}         & \multicolumn{1}{c|}{0.9885}        & 17.64            \\
                                                                                  & 4                                                                                    & \multicolumn{1}{c|}{33.30}         & \multicolumn{1}{c|}{0.9911}        & 18.25            \\ \hline
\end{tabular}}
\vspace{-10pt} 
\end{table}

\subsection{Attribution Analysis}
\vspace{-3pt}
Figure \ref{fig:lam} presents the results of the attribution analysis conducted using the proposed DAM. As illustrated in Figure \ref{fig:lam}, methods employing the slicing inference strategy are limited to local information in the vicinity of the attribution regions during the image reconstruction process. This restriction can lead to color distortions and artifacts, particularly in areas with complex textures or uneven brightness, thereby adversely affecting global consistency of dehazed results. In contrast, both 4KDehazing and the proposed DehazeXL can directly infer high-resolution images without the need for slicing strategies, allowing them to leverage global information to aid in the reconstruction of local areas, thus achieving better global consistency. Furthermore, compared to 4KDehazing, DehazeXL demonstrates a more efficient utilization of local features and global context, resulting in higher quality detail recovery and improved dehazing performance.

Additionally, the attribution maps shown in Figure \ref{fig:lam} (1-l) and (2-l) indicate that the model tends to focus on haze-free regions and high-contrast textures during the reconstruction process. This phenomenon suggests that the model prioritizes the use of unambiguous visual cues to enhance the quality of the dehazed output. Compared to methods employing slicing inference strategies, the proposed approach more effectively utilizes the spectral and color information from haze-free regions, thereby underscoring the importance of contextual information in efficient image dehazing.

 \vspace{-2pt}
\section{Conclusion}
\vspace{-2pt}
In this paper, we propose DehazeXL, an end-to-end haze removal method that effectively integrates global information with local feature, enabling efficient processing of large images while minimizing GPU memory usage. To facilitate a visual interpretation of the factors influencing dehazed results, we design the Dehazing Attribution Map for haze removal tasks. Quantitative and qualitative evaluations demonstrate that the proposed DehazeXL outperforms state-of-the-art haze removal techniques in terms of both accuracy and inference speed across multiple high-resolution datasets. The results of the attribution analysis underscore the critical role of global information in image dehazing tasks. Moreover, our work provides a valuable dataset (\emph{8KDehaze}) and analytical tool for future research in the field of haze removal.
 {
     \small
     \bibliographystyle{ieeenat_fullname}
     \bibliography{main}
 }


\end{document}